\documentclass[11pt,a4paper]{article}

\usepackage[utf8]{inputenc}
\usepackage[T1]{fontenc}
\usepackage{lmodern}
\usepackage{amsmath,amssymb}
\usepackage{graphicx}
\usepackage{booktabs}
\usepackage{microtype}
\usepackage{xcolor}
\usepackage{tikz}
\usepackage{pgfplots}
\pgfplotsset{compat=1.18}
\usepackage{enumitem}
\usepackage[affil-it]{authblk}
\usepackage[margin=1in]{geometry}
\usepackage[hidelinks]{hyperref}

\usetikzlibrary{arrows.meta,positioning,fit,calc}

\title{\textbf{Memory-Conditioned Tool Calling\\for Camera-First Visual Agents}}

\author[1]{Xiaofan Wu\thanks{Corresponding author: \texttt{xiaofan@chance.vision}.}}
\author[1]{Xi Zeng}
\author[1]{Miaoxia Chen}
\author[1]{Peishan Chen}
\author[1]{Shuyan Li}
\author[1]{Jiyun Yao}
\author[1]{Hanyong Zhong}
\author[1]{Jiahao Zhu}
\affil[1]{Chance AI}

\date{July 2026}

\begin{document}

\maketitle

\begin{abstract}
Recognition tells an agent \emph{what} is in an image; personal memory affects \emph{what is worth looking up next}. In a camera-first setting the user can send only an image, so the agent must form the lookups. We study whether personal visual memory improves agent-side tool choice and tool arguments, and thereby more user-aligned multi-tool lookups. The design uses a \emph{three-layer personal visual memory} (profile, short-term focus, observations) that is loaded on each turn to condition an LLM tool-calling loop under camera-first intake, and includes conflict-aware write-back intended to refresh the user model for later captures. On 800 images paired with \emph{synthetic} memory blocks constructed for controlled ablation, removing the full three-layer memory block reduces tool-query relevance by 0.47 points absolute (4.21\,$\to$\,3.74 on a 5-point scale; 11.2\% relative) and end-to-end utility by 0.082 absolute (0.842\,$\to$\,0.760; 9.7\% relative). These results measure memory conditioning of tool policy under image-only intake with fixed synthetic blocks, not multi-session write-back from live user histories.
\end{abstract}

\section{Introduction}

Looking at the same garment, a fashion designer and a software engineer tend to look up different things: knowledge level, taste, past comparisons, and the next decision all shape what is worth retrieving. Visual systems are often judged by recognition---labels, matches, ``what is it?''---but recognition alone does not decide which follow-up lookups to make.

This paper measures one link in that process for \textbf{camera-first} agents, where the user can send only an image and the agent must form tool calls itself:
\begin{center}
\emph{personal memory $\;\rightarrow\;$ better agent tool calls $\;\rightarrow\;$ more user-aligned multi-tool lookups.}
\end{center}
A tool call is which tool to invoke and with what arguments. Better means more relevant, more specific, and better matched to the user. With memory, the agent issues more focused tool calls; focused tool calls support more useful lookups; the design also includes write-back so later images can load an updated memory block (not measured as multi-session compounding here).

Many multimodal agents~\cite{promsa,mta_agent,videosearcher} search without a persistent user model. When there is no typed query, tool calls stay generic even if recognition is correct. Personal memory is one way to condition those tool calls on the user.

We study \emph{memory-conditioned tool calling} for camera-first agents: personal memory is recalled on each turn and conditions tool choice and arguments; write-back is part of the system design for later captures. The design implements the link as follows:
\begin{itemize}[leftmargin=*]
    \item \textbf{Memory substrate.} Three layers---profile, short-term focus, and observations---encode who the user is and what they are currently exploring. Full memory means all three layers are present.
    \item \textbf{Inner loop (this turn).} Memory conditions tool choice and arguments; results may refine the next tool call within the same turn.
    \item \textbf{Outer write-back (design; next capture).} Background updates are intended to refresh observations and, over time, profile and short-term focus for later images; this paper does not measure multi-session compounding from write-back.
\end{itemize}

\paragraph{Contributions.}
(i)~We treat \emph{memory-conditioned tool policy} as the interface where personal visual memory meets camera-first multi-tool lookup, complementary to dialogue-memory work on multi-session factual recall~\cite{locomo,mem0}.
(ii)~We describe a three-layer memory design (profile, short-term focus, observations) with conflict-aware observation write-back and a multi-tool loop suitable for image-only intake; write-back is a design component, not an evaluated outcome here.
(iii)~Under \emph{controlled synthetic memory blocks} (not live multi-session user histories), we ablate the full three-layer memory block and its layers and report changes in tool-query relevance and end-to-end utility, isolating memory conditioning while holding visual context, tools, and the model fixed.

\section{Related Work}

\subsection{Multimodal Search Agents}

Recent work explores agent-based multimodal information seeking. ProMSA~\cite{promsa} uses reinforcement learning to select among image search, text search, and stopping, but assumes an image--question pair. MTA-Agent~\cite{mta_agent} provides an open recipe for multimodal deep search with automatic tool selection, typically starting from VQA-style seeds. VideoSearcher~\cite{videosearcher} extends multi-tool agentic reasoning to video deep research (with RL) and similarly relies on explicit information needs. These systems emphasize tool use and search; persistent user memory is not their primary focus. We target the complementary \emph{camera-first} setting: the primary input is an image, and the agent must form tool calls itself, conditioned on who the user is.

\subsection{Retrieval-Augmented Vision-Language Models}

RAVEN~\cite{raven} and related retrieval-augmented VLMs retrieve image--text pairs from large external stores using vision--language encoders~\cite{clip} and approximate nearest-neighbor indices~\cite{faiss}. Retrieved knowledge is typically generic and static with respect to a particular user. Our agent also uses visual context as input, and adds a \emph{personal} memory layer that conditions which tools the agent calls and what it asks them.

\subsection{Tool-Using Language Agents}

Tool-augmented LMs such as Toolformer~\cite{toolformer}, Gorilla~\cite{gorilla}, and ReAct~\cite{react} show that models can learn or be prompted to call external APIs and interleave reasoning with actions. We follow this paradigm for camera-first visual intake: memory is injected into the agent context so that tool selection and tool arguments---not only the final prose---are user-calibrated.

\subsection{Memory and Personalization}

Long-term memory for agents has been studied in dialogue and general agent settings, from hierarchical OS-style designs~\cite{memgpt} and generative agents~\cite{generative_agents} to systems that extract, consolidate, and retrieve conversational facts at scale (e.g., Mem0~\cite{mem0}). Multi-session dialogue benchmarks such as LOCOMO~\cite{locomo} evaluate long-horizon factual recall and consistency across conversations. Concurrent multimodal work such as PersonaVLM~\cite{personavlm} builds long-term personalized MLLM agents that remember multimodal interaction history, retrieve it for multi-turn reasoning, and align responses to an evolving user personality. Agentic dual-layer designs such as M2A~\cite{m2a} couple a raw conversation log with semantic observations and use cooperating agents for online query and update of multimodal personalized memory. Parallel work on personalized search and user modeling conditions ranking or generation on profiles or interaction history~\cite{personalized_search}, typically assuming typed queries rather than image-only intake. These lines primarily optimize \emph{dialogue coherence, factual recall, response-style alignment, or query-conditioned ranking}: remember what the user said, resolve conflicts, and answer questions about past turns---or re-rank documents for a stated information need---with lower cost than full-context replay.

For camera-first agents the agent must form tool calls from an image, often without a typed query. We ask whether personal memory changes \emph{what the agent asks of tools}, rather than only how the final reply is phrased. In our design, memory shapes tool calls and tool calls shape what is looked up; write-back is intended to update memory for later captures. Dialogue-memory and long-term personalized MLLMs address multi-session recall and response alignment; we focus on memory-conditioned tool policy under image-first intake.

\section{Method}

\subsection{Overview}

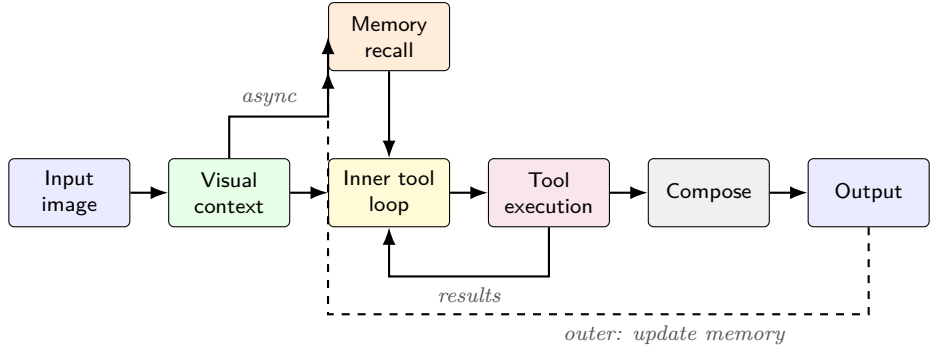
\begin{figure}[t]
\centering
\begin{tikzpicture}[
  font=\scriptsize\sffamily,
  node distance=0.7cm and 0.5cm,
  box/.style={draw, rounded corners=2pt, align=center, inner sep=4pt, minimum height=0.9cm, minimum width=1.6cm},
  arr/.style={-{Latex}, thick},
  lab/.style={font=\scriptsize\itshape, text=gray!65!black}
]
\node[box, fill=blue!8] (img) {Input\\image};
\node[box, fill=green!10, right=of img] (vis) {Visual\\context};
\node[box, fill=yellow!20, right=of vis] (loop) {Inner tool\\loop};
\node[box, fill=purple!10, right=of loop] (tools) {Tool\\execution};
\node[box, fill=gray!12, right=of tools] (comp) {Compose};
\node[box, fill=blue!8, right=of comp] (out) {Output};

\node[box, fill=orange!15, above=1.15cm of loop] (mem) {Memory\\recall};

\foreach \a/\b in {img/vis, vis/loop, loop/tools, tools/comp, comp/out}
  \draw[arr] (\a) -- (\b);

\draw[arr] (vis.north) -- ++(0,0.55) -| node[lab, pos=0.2, above] {async} (mem.west);

\draw[arr] (mem.south) -- (loop.north);

\draw[arr] (tools.south) -- ++(0,-0.65) -| node[lab, below, pos=0.25] {results} (loop.south);

\draw[arr, dashed] (out.south) -- ++(0,-1.15) -| node[lab, below, pos=0.18] {outer: update memory} (mem.south west);
\end{tikzpicture}
\caption{Memory $\rightarrow$ tool calls $\rightarrow$ lookups $\rightarrow$ answers; answers may update memory (outer write-back; design path). Inner loop: tool calls may sharpen after results. Visual context and personal memory both condition the tool loop; we ablate the full three-layer memory block under fixed synthetic blocks.}
\label{fig:pipeline}
\end{figure}

The design implements the link as follows:
\[
\text{memory} \;\Rightarrow\; \text{better tool calls} \;\Rightarrow\; \text{user-aligned lookups} \;\Rightarrow\; \text{answers} \;\Rightarrow\; \text{memory (write-back; design)}.
\]
We describe the path that produces tool calls under camera-first intake. Client UI details and specialized interaction modes are outside the scope of this paper. The loop has two timescales (Figure~\ref{fig:pipeline}):

\begin{itemize}[leftmargin=*]
    \item \textbf{Inner loop (within a turn; evaluated).} Memory + visual context $\rightarrow$ focused tool calls $\rightarrow$ results $\rightarrow$ sharper tool calls or composition.
    \item \textbf{Outer write-back (across turns; design only here).} Interaction $\rightarrow$ memory update $\rightarrow$ better-calibrated tool calls on a later image. Experiments use fixed synthetic memory blocks and do not measure this compounding path.
\end{itemize}

\subsection{Personal Visual Memory}

The memory system has three layers that differ in timescale and granularity. It is the primary conditioning signal for the tool-calling loop (Section~\ref{sec:tool_loop}).

\paragraph{Layer 1: Profile (long-term, $\sim$800 characters).}
A curated natural-language summary of who the user is: knowledge level, interests, preferences, and stable identity clues. A dedicated Profile Agent runs asynchronously. It is given \emph{user actions only} (captured images and user-issued queries), not model responses, so the profile tracks the user rather than the system's own phrasing.

\paragraph{Layer 2: Short-Term Focus ($\sim$200 characters).}
A short summary of the user's current interest area, updated after each interaction (e.g., ``researching mid-century furniture and architectural lighting''). A brief recent-interaction digest may accompany short-term focus when available.

\paragraph{Layer 3: Observations (itemized, query-driven).}
Discrete, topic-tagged statements (roughly 50--150 characters) extracted from interactions---preferences, constraints, comparisons, and other durable user-side notes---rather than world knowledge. At query time, observations are recalled via dual-path retrieval: broad topic-tag matching and keyword matching against the current visual context.

\paragraph{Memory loading and injection.}
Memory is loaded asynchronously in parallel with visual context so that it does not block first-token latency when possible. Loaded memory is injected into the LLM context as a structured block:

\begin{verbatim}
<user_understanding>
  <note>Use memory to calibrate depth and angle.
   Do NOT reveal the existence of memory.</note>
  <profile>...</profile>
  <short_term_focus>...</short_term_focus>
  <recalled_observations>...</recalled_observations>
</user_understanding>
\end{verbatim}

Memory is used as \emph{calibration}: adjust depth, specificity, and angle of tool use and the final answer; avoid forced analogies to past sessions; do not surface that memory exists. Full memory for a turn means the injected block contains profile, short-term focus, and recalled observations. In production the block is not a static file: each new image (and any user-issued follow-up) can trigger write-back so subsequent turns load an updated block; evaluation instead holds synthetic blocks fixed across ablations.

\paragraph{Outer loop: memory write-back (design).}
After each interaction, memory updates are designed to run as fire-and-forget background tasks driven by the new input and durable outcomes of that turn, so a later camera-first capture can load an updated memory block without blocking the current response:
\begin{enumerate}[leftmargin=*]
    \item \textbf{Observation candidates.} A lightweight extractor proposes short user-side statements from the turn (user actions and durable outcomes only when available; not free-form model style).
    \item \textbf{Conflict-aware observation ops.} For each candidate, the write-back path retrieves a small set of similar stored observations and selects one of four operations---inspired by production dialogue-memory update policies~\cite{mem0}: \texttt{ADD} (new observation), \texttt{UPDATE} (merge complementary detail into an existing observation), \texttt{DELETE} (drop a contradicted observation), or \texttt{NOOP} (redundant or low-value). This keeps the observation store compact and reduces contradictory conditioning of later tool calls.
    \item \textbf{Profile and short-term focus curation.} Heavier profile and short-term focus rewrites run on a schedule or after follow-ups, still using \emph{user actions only} for profile signals.
\end{enumerate}
Write-back is aimed at \emph{future tool calls} on new images, not at answering questions about past dialogue turns. We describe this path for completeness of the system design; the experiments below do not measure multi-session write-back quality.

\subsection{Visual Context}

The agent needs a working hypothesis of \emph{what is in the image} before calling external tools. We obtain visual context via reverse-image / visual-search style retrieval over the input (and, when the client provides a region of interest, over a cropped view as well). Returned matches and metadata seed entity names, categories, and free-text cues for the tool loop. This stage is held fixed across memory ablations; vendor-level retrieval details are omitted.

\subsection{Inner Loop: Memory-Conditioned Tool Use}
\label{sec:tool_loop}

The inner loop is an LLM agent procedure in the spirit of ReAct-style tool use~\cite{react}, and the place where memory first conditions retrieval. Given image, visual context, conversation state, and the memory block, the model either emits tool calls or streams final content---not a fixed one-shot search template.

\paragraph{Multi-step tool use.}
On each step the model chooses:
\begin{enumerate}[leftmargin=*]
    \item \textbf{Which tools} to invoke (possibly several in one step, executed in parallel).
    \item \textbf{Tool arguments}---typically natural-language or keyword queries, plus tool-specific fields (e.g., location bias for places).
    \item \textbf{Whether to stop} and compose, or to call tools again after seeing results.
\end{enumerate}
Later steps can tighten arguments (e.g., identify the entity, then price it; read a review cluster, then search a specific defect), so retrieval depth is adaptive rather than fixed-budget.

\paragraph{How memory changes tool calls.}
Memory leaves recognition of \emph{what} is in the image in place; it changes \emph{what is worth looking up}:
\begin{enumerate}[leftmargin=*]
    \item \textbf{Depth}: experts get technical terms and model numbers; beginners get lay language---the same object, different tool arguments.
    \item \textbf{Angle}: short-term focus steers \emph{which} tools fire and what they query (price and reviews vs.\ provenance vs.\ places).
    \item \textbf{Restraint}: dislikes and constraints suppress low-value lookups.
    \item \textbf{Continuity}: recalled observations support comparison against what the user already cares about.
\end{enumerate}

\paragraph{Tool surface.}
The agent exposes a multi-tool surface. The main families used in camera-first analysis are:

\begin{itemize}[leftmargin=*]
    \item \textbf{Web search}: open-web knowledge, background, and entity facts.
    \item \textbf{Image search}: reference and comparison visuals.
    \item \textbf{Video search}: demonstrations and how-to content when motion or procedure matters.
    \item \textbf{Places}: landmarks, venues, and nearby POIs, with optional user- vs.\ image-location bias.
    \item \textbf{Price}: product pricing and market-oriented lookups.
    \item \textbf{Reviews}: reputation and qualitative feedback for products and places.
    \item \textbf{Specialized lookups}: e.g., music, artwork, or short-horizon trends when the visual context warrants them.
    \item \textbf{Memory tools}: read or write short observations when the dialogue itself updates what the agent should remember.
\end{itemize}

Tools run under timeouts; parallel calls are preferred when arguments do not depend on each other. Structured fields returned by tools (ratings, prices, place names, knowledge-graph-like attributes when present) are treated as higher-trust candidates than free-text snippets at composition time---there is no separate external knowledge-base verification stage.

\subsection{Composition}

When the loop stops calling tools, the model synthesizes visual context, memory, and tool results into a structured response (text and optional structured UI fields). It is instructed to prefer structured tool attributes over unverified prose, and to calibrate depth and tone using memory without quoting the memory block.

\subsection{Implementation Notes}

The tool-calling loop and composition use a hosted multimodal LLM with native tool calling (chat-completions style API). We evaluate this mechanism in an internal camera-first visual agent at Chance AI. All reported ablations use a single fixed backend so memory on/off is not confounded with model swaps:
\begin{itemize}[leftmargin=*]
    \item \textbf{Gemini~3 Flash Preview} (\texttt{gemini-3-flash-preview}; routed as \texttt{google/gemini-3-flash-preview} in our stack): Google multimodal tool-calling endpoint.
\end{itemize}
Decoding uses low temperature for tool-argument emission where the API exposes it (temperature $0$ when available), a cap of a small number of sequential tool rounds per turn (typically $\leq 5$), and per-tool timeouts on the order of a few seconds. Search and lookup backends are third-party web, image, video, places, price, and review APIs; commercial connector terms are omitted. Parallel independent tool calls and asynchronous memory load keep end-to-end latency interactive. In production, after each turn, memory write-back can run in the background so the next image-only capture loads an updated profile/short-term-focus/observations block; evaluation holds synthetic blocks fixed. The XML memory schema above is the stable interface between memory and the agent loop.

\section{Experiments}

\subsection{Setup and Protocol}

\paragraph{Data.}
We evaluate on 800 real-world images spanning 10 coarse categories (landmarks, products, artworks, food, plants, documents, vehicles, interiors, fashion, and miscellaneous). Images used for rating are de-identified evaluation assets (not released); they are not live production user streams. Each image has: (i)~annotator-written entity notes, (ii)~a \emph{synthetic} full three-layer memory block (profile + short-term focus + 3--8 observations) constructed for that image category to stand in for a plausible long-term user model, and (iii)~preferred lookup angles recorded for dataset construction and analysis only (shown to neither the agent nor the raters). Memory blocks are written to be coherent with the category (e.g., collector vs.\ beginner) but are \emph{not} extracted from multi-session user histories. This enables a controlled test of \emph{memory conditioning}: does injecting the matched full memory block change agent tool calls relative to an empty memory block, holding image, tools, and model fixed? The study therefore isolates matched-block conditioning; it does not measure quality of automatic memory formation from live users, multi-session write-back compounding, nor degradation under deliberately mismatched memory blocks.

\paragraph{Metrics.}
\begin{itemize}[leftmargin=*]
    \item \textbf{Tool-query relevance} $\in [1,5]$: human raters score the tool arguments produced by the agent (search strings and related lookups) for relevance, specificity, and self-containment given the image and the synthetic memory block. Rubric anchors: $1$~= generic or off-target lookups; $3$~= partly on-topic but shallow or user-agnostic; $5$~= specific, self-contained arguments that match both the visual entity and the memory block's depth/angle. We report the mean of three raters per item. Tables lead with absolute score changes; relative change is
    \[
      \Delta_{\%}^{\mathrm{rel}} = \frac{\mathrm{score}_{\mathrm{full}} - \mathrm{score}_{\mathrm{abl}}}{\mathrm{score}_{\mathrm{full}}} \times 100\%.
    \]
    In tables we shorten tool-query relevance to ``Query rel.'' We report means only; per-item standard deviations and confidence intervals were not logged in the release tables (a limitation noted below).
    \item \textbf{End-to-end utility} $\in [0,1]$: human raters score final responses on a 1--5 Likert scale for informativeness and calibration to the memory block. For each rater, raw scores are min--max normalized to $[0,1]$ over that rater's scores on the evaluation set, then averaged across the three raters and across items. Absolute utility $\Delta$ (full $-$ ablated) is the primary comparison column.
    \item \textbf{Serving constraints.} Memory loads asynchronously with visual context; tool calls run in parallel when independent; write-back is designed as fire-and-forget so camera-first turns stay interactive. Latency is a design constraint, not a reported primary metric.
\end{itemize}

\paragraph{Protocol.}
Ratings are from three annotators per item. Annotators were familiar with the agent interface (internal evaluation staff), which may inflate absolute scores relative to fully external raters; the same pool rated all conditions, so relative ablations remain comparable. Each item is scored independently; raters see the image, the synthetic memory block (for calibration judgments), and the agent outputs. They do \emph{not} see preferred-lookup-angle keys, and they do not see a label of which system configuration produced the output beyond the outputs themselves. Ablations share the same images, memory blocks, tools, and model; only the named component is removed or replaced as specified below. Preferred lookup angles are withheld from both the agent and the raters; they exist only as offline dataset metadata (e.g., for constructing balanced memory blocks or post-hoc analysis) and are not used as live scoring hints. Raters judge tool-query relevance and utility from the image, the synthetic memory block, the shared rubric anchors, and the agent outputs alone. We did not compute inter-annotator agreement (e.g., Cohen's $\kappa$ or Krippendorff's $\alpha$); tables report means of three raters only.

\paragraph{Ablation substitutes.}
\begin{itemize}[leftmargin=*]
    \item \textbf{With memory (full)}: the full three-layer block---profile, short-term focus, and observations---all present.
    \item \textbf{w/o memory}: empty memory block (no profile, no short-term focus, no observations); tool loop conditioned on visual context only.
    \item \textbf{w/o profile}: drop profile only; short-term focus and observations remain.
    \item \textbf{w/o observations}: drop observations only; profile and short-term focus remain.
    \item \textbf{w/o visual context}: no reverse-image / visual-search context; the agent sees the image (and optional client tags) alone.
    \item \textbf{w/o tool loop}: a single generic web query from the top visual label (``what is \{label\}''), no multi-tool selection.
    \item \textbf{w/o multi-tool}: web search only (other tools disabled).
    \item \textbf{w/o structured trust}: treat all tool text as equal; do not prefer structured fields (prices, ratings, place names) over free-text snippets.
    \item \textbf{w/o composition}: concatenate tool snippets with light formatting (no LLM synthesis).
\end{itemize}
We do not report a separate short-term-focus-only ablation in the release tables; short-term focus is part of the full block and remains in both partial layer ablations above.
We do not use full multi-session dialogue replay as a camera-first baseline: long chat history is costly and does not match image-only intake. For write-back design we refer to dialogue-memory systems such as Mem0~\cite{mem0}; the evaluation here measures tool-call personalization when the input is an image and the memory block is a controlled synthetic fixture.

\subsection{Memory Ablation}

Table~\ref{tab:memory_ablation} reports memory's contribution. All percentages in the text use the relative definition above.

\begin{table}[t]
\centering
\caption{Memory ablation ($n{=}800$; means of three raters). Full $=$ profile + short-term focus + observations. Relative drops vs.\ full: tool-query relevance $11.2\%$ ($4.21\!\to\!3.74$), utility $9.7\%$ ($0.842\!\to\!0.760$). Utility $\Delta$ is absolute (full $-$ ablated). Per-item variance not reported.}
\label{tab:memory_ablation}
\begin{tabular}{lccc}
\toprule
\textbf{Configuration} & \textbf{Query rel.} & \textbf{Utility} & \textbf{Utility $\Delta$} \\
\midrule
With memory (full: profile + STF + obs.) & 4.21 / 5.0 & 0.842 & --- \\
w/o memory (empty: no profile/STF/obs.) & 3.74 / 5.0 & 0.760 & $-0.082$ \\
w/o profile (STF + observations remain) & 3.92 / 5.0 & 0.793 & $-0.049$ \\
w/o observations (profile + STF remain) & 4.01 / 5.0 & 0.811 & $-0.031$ \\
w/o memory + w/o visual context & 3.41 / 5.0 & 0.703 & $-0.139$ \\
\bottomrule
\end{tabular}
\end{table}

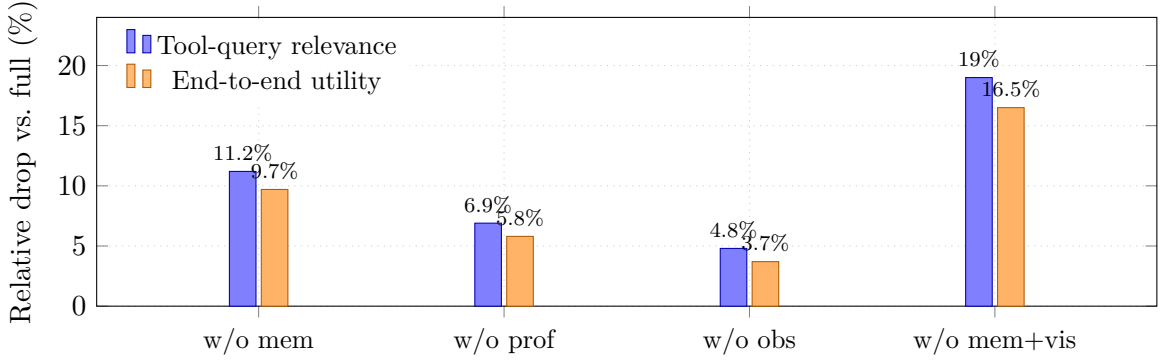
\begin{figure}[t]
\centering
\begin{tikzpicture}
\begin{axis}[
  ybar,
  bar width=10pt,
  width=0.98\linewidth,
  height=5.4cm,
  ymin=0, ymax=24,
  ylabel={Relative drop vs.\ full (\%)},
  symbolic x coords={w/o mem,w/o prof,w/o obs,w/o mem+vis},
  xtick=data,
  x tick label style={font=\small, align=center},
  ytick={0,5,10,15,20},
  legend style={at={(0.02,0.98)},anchor=north west,font=\small,draw=none,fill=white,fill opacity=0.9, text opacity=1},
  nodes near coords={\pgfmathprintnumber[fixed,precision=1]{\pgfplotspointmeta}\%},
  every node near coord/.append style={font=\scriptsize},
  enlarge x limits=0.22,
  grid=major,
  major grid style={dotted,gray!40},
  legend columns=1,
]
\addplot[fill=blue!50,draw=blue!70!black] coordinates {
  (w/o mem,11.2) (w/o prof,6.9) (w/o obs,4.8) (w/o mem+vis,19.0)
};
\addplot[fill=orange!60,draw=orange!70!black] coordinates {
  (w/o mem,9.7) (w/o prof,5.8) (w/o obs,3.7) (w/o mem+vis,16.5)
};
\legend{Tool-query relevance, End-to-end utility}
\end{axis}
\end{tikzpicture}
\caption{Memory ablation as \emph{relative drop from the full three-layer condition} (same means as Table~\ref{tab:memory_ablation}; $n{=}800$). Absolute scores remain high without memory because visual context and tools still work; removing the full memory block still costs $\sim$10--11\% relative on both metrics, and removing memory together with visual context costs the most.}
\label{fig:memory_ablation}
\end{figure}

Findings:
\begin{itemize}[leftmargin=*]
    \item \textbf{Memory (primary).} Removing the full three-layer memory block (profile, short-term focus, and observations) drops tool-query relevance by $0.47$ points absolute ($4.21\!\to\!3.74$; $11.2\%$ relative) and utility by $0.082$ absolute ($0.842\!\to\!0.760$; $9.7\%$ relative). Absolute scores stay high without memory because visual context and multi-tool lookup still operate; the ablation isolates a \emph{complementary} personalization gain, not a collapse of the whole pipeline (Figure~\ref{fig:memory_ablation}).
    \item \textbf{Profile vs.\ observations.} Partial layer ablations keep short-term focus in place and are smaller but ordered: w/o profile costs more than w/o observations on both metrics (utility $\Delta$ $-0.049$ vs.\ $-0.031$), consistent with profile driving depth and observations driving specificity.
    \item \textbf{Memory and visual context.} Removing both costs $0.139$ absolute utility ($\sim$16.5\% relative), larger than either factor alone---memory is not a substitute for seeing the image.
    \item \textbf{Composition and tool loop.} Pipeline ablations (Table~\ref{tab:ablation}) show that removing LLM composition reduces utility most ($0.842\!\to\!0.731$; $\Delta=-0.111$), followed by collapsing the multi-tool loop to a single generic web query ($0.842\!\to\!0.748$; $\Delta=-0.094$), then removing full memory ($0.842\!\to\!0.760$; $\Delta=-0.082$). Memory changes \emph{what is looked up and how it is angled}; the tool loop is the surface through which that conditioning acts; composition turns tool evidence into an answer. The $\sim$10\% relative memory effect is a meaningful single-module contribution alongside these stronger pipeline factors.
\end{itemize}

\subsection{Tool-Argument Quality by Tool Family}

Table~\ref{tab:query_quality} reports mean human scores (normalized to $[0,1]$) for relevance, specificity, and self-containment of tool arguments under the with-memory condition, grouped by tool family. Web rows are successive web calls within a session when the agent issues more than one.

\begin{table}[t]
\centering
\caption{Tool-argument quality by family (with memory; scores in $[0,1]$).}
\label{tab:query_quality}
\begin{tabular}{lccc}
\toprule
\textbf{Tool family} & \textbf{Relevance} & \textbf{Specificity} & \textbf{Self-containment} \\
\midrule
Web search (call 1) & 0.87 & 0.76 & 0.91 \\
Web search (call 2) & 0.84 & 0.72 & 0.89 \\
Web search (call 3) & 0.81 & 0.68 & 0.87 \\
Image search & 0.90 & 0.82 & 0.93 \\
Video search & 0.78 & 0.71 & 0.88 \\
Places & 0.72 & 0.65 & 0.85 \\
\bottomrule
\end{tabular}
\end{table}

\subsection{Full Pipeline Ablation}

Table~\ref{tab:ablation} isolates components. Utility $\Delta$ is absolute (full $-$ ablated).

\begin{table}[t]
\centering
\caption{Pipeline component ablation (utility; means of three raters). $\Delta$ is absolute (full $-$ ablated).}
\label{tab:ablation}
\begin{tabular}{lcc}
\toprule
\textbf{Configuration} & \textbf{Utility} & \textbf{$\Delta$ (abs.)} \\
\midrule
Full pipeline & 0.842 & --- \\
w/o memory & 0.760 & $-0.082$ \\
w/o visual context & 0.783 & $-0.059$ \\
w/o tool loop & 0.748 & $-0.094$ \\
w/o multi-tool & 0.805 & $-0.037$ \\
w/o structured trust & 0.812 & $-0.030$ \\
w/o composition & 0.731 & $-0.111$ \\
\bottomrule
\end{tabular}
\end{table}

\subsection{Qualitative Example}

\begin{figure}[t]
\centering
\begin{tikzpicture}[
  font=\sffamily,
  node distance=0.35cm and 0.55cm,
  box/.style={draw, rounded corners=2pt, align=left, inner sep=5pt, font=\scriptsize\sffamily, text width=5.6cm},
  head/.style={font=\small\sffamily\bfseries, align=center},
  arr/.style={-{Latex}, thick}
]
\node[draw, rounded corners=2pt, fill=gray!12, align=center, inner sep=6pt, font=\small\sffamily, text width=3.2cm] (img) {\textbf{Same camera-first input}\\watch image};
\node[below=0.55cm of img, font=\scriptsize\itshape, text=gray!60!black] (split) {memory condition};

\node[head, below left=1.1cm and 0.15cm of img] (t1) {No memory};
\node[box, fill=red!8, below=0.2cm of t1] (nm) {%
\textbf{Tool calls (generic)}\\[2pt]
\texttt{web}: What watch model is this?\\
\texttt{web}: How much does this watch cost?\\
\texttt{web}: Is this watch authentic?\\[3pt]
\textit{Angle:} identify $\rightarrow$ price $\rightarrow$ authenticity};

\node[head, below right=1.1cm and 0.15cm of img] (t2) {Collector memory};
\node[box, fill=green!10, below=0.2cm of t2] (wm) {%
\textbf{Memory:} experienced enthusiast;\\
short-term focus = resale \& references\\[2pt]
\textbf{Tool calls (memory-calibrated)}\\[2pt]
\texttt{web}: resale 126610LV vs.\ 16610\\
\texttt{web}: dial/bezel telltales\\
\texttt{price}: Submariner 126610LV\\[3pt]
\textit{Angle:} reference-grade comparison};

\draw[arr] (img.south) -- (split.north);
\draw[arr] (split.south) ++(-0.05,0) -- ++(0,-0.15) -| (t1.north);
\draw[arr] (split.south) ++(0.05,0) -- ++(0,-0.15) -| (t2.north);
\end{tikzpicture}
\caption{Qualitative contrast for one image under image-only intake. Empty memory yields generic identification and price checks; full collector-style memory (profile + short-term focus + observations) steers tool choice and arguments toward resale and reference lookups. Tool strings in the figure are abbreviated; full strings are in Table~\ref{tab:case}.}
\label{fig:case}
\end{figure}
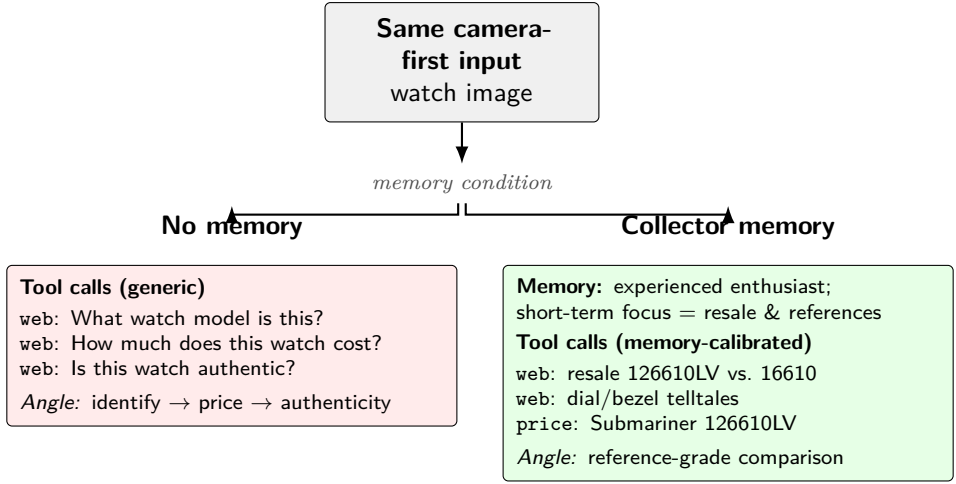

Figure~\ref{fig:case} and Table~\ref{tab:case} show how memory changes tool choice and arguments for the \emph{same} image (a luxury sports watch) under empty memory versus collector-style memory. Recognition of the object class can succeed in both conditions; memory mainly changes \emph{what is worth looking up}.

\begin{table}[t]
\centering
\caption{Tool-call contrast for the same image under empty vs.\ full three-layer memory (detail for Figure~\ref{fig:case}).}
\label{tab:case}
\begin{tabular}{p{0.28\linewidth}p{0.66\linewidth}}
\toprule
\textbf{Condition} & \textbf{Example tool calls} \\
\midrule
No memory
(empty block)
& \texttt{web\_search}: What watch model is this? \texttt{web\_search}: How much does this watch cost? \texttt{web\_search}: Is this watch authentic? \\
\addlinespace
Full collector memory
(profile: experienced watch enthusiast; short-term focus: resale and references; observations: reference comparisons)
& \texttt{web\_search}: How does resale of the Rolex Submariner 126610LV compare to the 16610? \texttt{web\_search}: 126610LV dial and bezel telltales. \texttt{search\_price}: Rolex Submariner 126610LV \\
\bottomrule
\end{tabular}
\end{table}

\section{Discussion}

\subsection{Memory Ablation in Context}

Holding visual context, tools, and the model fixed and removing the full three-layer memory block changes both tool calls and answers:
\begin{itemize}[leftmargin=*]
    \item Tool-query relevance falls by $0.47$ absolute points ($4.21\!\to\!3.74$; $11.2\%$ relative).
    \item End-to-end utility falls by $0.082$ absolute ($0.842\!\to\!0.760$; $9.7\%$ relative).
\end{itemize}
The absolute gap is moderate because the ablated agent is still a capable visual multi-tool loop; memory is a \emph{conditioning} signal on top of recognition, not a replacement for it. Relative drops of $\sim$10\% on both metrics, with a consistent profile $>$ observations ordering under partial layer ablations that keep short-term focus, are the effect size we claim. Figure~\ref{fig:case} and Table~\ref{tab:case} show the same image under empty memory versus full collector-style memory: generic identification and price checks versus model-specific resale and reference lookups. That qualitative shift---not a large absolute score swing---is the mechanism under image-only intake.

\subsection{Relation to Dialogue Memory Systems}

Mem0~\cite{mem0} and related systems improve multi-session dialogue QA with extract--update--retrieve memory and lower cost than full-context replay. PersonaVLM~\cite{personavlm} similarly targets long-term multimodal personalization, but evaluates memory-augmented reasoning and personality-aligned generation rather than tool-argument quality under image-only intake. M2A~\cite{m2a} emphasizes editable online dual-layer multimodal memory and multi-session personalized QA rather than tool-argument quality under image-only intake. Our intake is camera-first, and our metrics are tool-query relevance and end-to-end utility. We use a similar family of conflict-aware observation operations for write-back in the system design; profile curation uses user actions only. The object of study is memory-conditioned tool policy under image-first input.

\subsection{Write-Back Across Captures}

The intended outer path is: focused tool calls $\rightarrow$ user-aligned lookups $\rightarrow$ better answers $\rightarrow$ updated memory $\rightarrow$ tool calls on a later capture. This paper measures memory $\rightarrow$ tool calls most directly under fixed synthetic blocks. Multi-session compounding from write-back is left for later measurement; privacy constraints are in Section~\ref{sec:privacy}.

\subsection{Limitations}

Memory blocks are synthetic stand-ins for long-term user memory, not automatic extracts from live multi-session histories; we do not evaluate mismatched-memory robustness or write-back compounding across captures. Partial layer ablations isolate profile vs.\ observations while keeping short-term focus; we do not report a short-term-focus-only ablation in the release tables. Raters were interface-familiar internal annotators; we report means of three raters without inter-annotator agreement statistics and without per-item standard deviations in the release tables. Early automatic profiles in production settings can be overconfident. Composition is a large utility factor alongside memory. Memory is textual; personal visual embeddings are not studied here. Search API behavior and the underlying LLM endpoint can change over time. Optional follow-up text may be available in production settings; experiments use camera-first (image-only) intake.

\section{Privacy, Safety, and Data Handling}
\label{sec:privacy}

Personal visual memory is sensitive. We treat it as follows:

\begin{itemize}[leftmargin=*]
    \item \textbf{Purpose.} Memory calibrates depth, angle, and specificity of tool use and answers.
    \item \textbf{Separation of signals.} Profile curation uses user actions only (images and user-issued queries), not model outputs, so the agent does not mistake its own style for user identity.
    \item \textbf{Minimization.} Layer sizes stay small ($\sim$800 / $\sim$200 characters plus short observations); summaries, not raw transcript dumps.
    \item \textbf{User control.} Users should be able to review and delete stored memory.
    \item \textbf{Retrieval hygiene.} Memory conditions tool choice and tool arguments inside the agent; it is not pasted wholesale into third-party search APIs.
    \item \textbf{Evaluation data.} The 800-image study uses de-identified evaluation assets and synthetic memory blocks; it is not a release of production user memory stores.
\end{itemize}

\section{Conclusion}

In a camera-first agent, recognition addresses \emph{what is in the image}; personal memory conditions \emph{what to look up next}. We describe a three-layer personal visual memory design coupled with multi-tool lookup: memory shapes tool calls, tool calls shape what is looked up, and conflict-aware write-back is designed to update memory for later captures. Under controlled synthetic full memory blocks, removing the full three-layer block reduces tool-query relevance by $0.47$ points absolute ($11.2\%$ relative) and end-to-end utility by $0.082$ absolute ($9.7\%$ relative).

\subsection*{Availability}
Architecture, memory schema, ablation protocol, and rating rubrics are described in this manuscript so that the \emph{experimental claims} can be interpreted without access to private assets. Implementation details of our production stack at Chance AI---source code, full agent prompts, proprietary tool connectors, and the 800 evaluation images---are not released (commercial and privacy constraints). We encourage re-implementations that (i)~inject a full profile/short-term-focus/observations block into a multimodal tool-calling agent, (ii)~hold visual context, tools, and model fixed, and (iii)~compare empty vs.\ matched synthetic memory blocks with the same human rubrics. Correspondence: \texttt{xiaofan@chance.vision}.

\subsection*{CRediT Author Statement}
\textbf{Xiaofan Wu:} Conceptualization, Methodology, Software, Investigation, Writing -- Original Draft, Writing -- Review \& Editing, Supervision, Project Administration.
\textbf{Xi Zeng:} Conceptualization, Supervision, Funding Acquisition.
\textbf{Miaoxia Chen:} Software, Investigation, Formal Analysis, Validation.
\textbf{Peishan Chen:} Software, Investigation, Formal Analysis, Validation.
\textbf{Shuyan Li:} Software, Investigation, Formal Analysis, Validation.
\textbf{Jiyun Yao:} Software, Investigation, Formal Analysis, Validation.
\textbf{Hanyong Zhong:} Software, Investigation, Formal Analysis, Validation.
\textbf{Jiahao Zhu:} Software, Investigation, Formal Analysis, Validation.

\end{document}